\documentclass[sigconf]{acmart}

\AtBeginDocument{%
  }

\setcopyright{acmcopyright}
\copyrightyear{2018}
\acmYear{2018}
\acmDOI{XXXXXXX.XXXXXXX}

\acmConference[Conference acronym 'XX]{Make sure to enter the correct
  conference title from your rights confirmation emai}{June 03--05,
  2018}{Woodstock, NY}
\acmPrice{15.00}
\acmISBN{978-1-4503-XXXX-X/18/06}

\usepackage{multirow}
\usepackage{subfigure}
\usepackage{algorithm}
\usepackage{algorithmic}
\usepackage{newfloat}
\usepackage{listings}
\usepackage{CJK}
\begin{document}

\title{MEDIC: A Multimodal Empathy Dataset in Counseling}

\author{Zhou'an Zhu, Xin Li, Jicai Pan, Yufei Xiao, Yanan Chang, Feiyi Zheng, Shangfei Wang}
\renewcommand{\shortauthors}{Trovato et al.}

\begin{abstract}
Although empathic interaction between counselor and client is fundamental to success in the psychotherapeutic process, there are currently few datasets to aid a computational approach to empathy understanding. In this paper, we construct a multimodal empathy dataset collected from face-to-face psychological counseling sessions. The dataset consists of 771 video clips. We also propose three labels (i.e., expression of experience, emotional reaction, and cognitive reaction) to describe the degree of empathy between the counselors and their clients.  Expression of experience describes whether the client has expressed experiences that can trigger empathy, and emotional and cognitive reactions indicate the counselor's empathic reactions. As an elementary assessment of the usability of the constructed multimodal empathy dataset, an interrater reliability analysis of annotators' subjective evaluations for video clips is conducted using the intraclass correlation coefficient and Fleiss' Kappa. Results prove that our data annotation is reliable. Furthermore, we conduct empathy prediction using three typical methods, including the tensor fusion network, the sentimental words aware fusion network, and a simple concatenation model. The experimental results show that empathy can be well predicted on our dataset. Our dataset is available for research purposes.
\end{abstract}

\maketitle

\section{Introduction}
In recent years, people have experienced more emotional distress from everyday life \cite{EmilyAHolmes2018TheLP}. Thus, they need emotional support, and often psychological treatment to recover. The counselors provide professional mental health support via psychological counseling  \cite{lambert1994effectiveness,mcleod2003doing}. Empathy is the basis of the humanistic approach to psychotherapy \cite{elliott2013research} and has long been used effectively. Empathy represents the therapist’s ability and willingness to understand the patient’s thoughts, feelings, and struggles from the patient’s perspective \cite{MatthewDavis1980AMA}. The ability to empathize is important in promoting positive behavior towards others, and may be the mechanism that drives the desire to help others \cite{AshishSharma2020ACA}. 

Although empathy has been studied extensively in the field of psychology \cite{barrett1981empathy,GodfreyTBarrettLennard2011ThePA,MatthewDavis1980AMA,FransBMdeWaal2008PuttingTA,elliott2011empathy,TaniaSinger2009TheSN}, the computational approach to understanding empathy during mental health support has been hindered by the lack of datasets. To the best of our knowledge, there is only one public dataset for empathy understanding \cite{AshishSharma2020ACA}, collected from text-based, asynchronous conversations on mental health platforms. Since psychological treatment frequently occurs in synchronous, face-to-face settings, a multimodal empathy dataset collected from face-to-face mental health support is urgently required. 

\begin{table*}[t]
\centering
\resizebox{.95\textwidth}{!}{
\setlength{\tabcolsep}{.8mm}
\begin{tabular}{l|l|l|l|l|l}
\hline
\multicolumn{1}{c|}{Dataset}                                      & \multicolumn{1}{c|}{Scenes}             & \multicolumn{1}{c|}{Annotation}                                                                                          & \multicolumn{1}{c|}{Modal}   & Data Num & \multicolumn{1}{c}{Public} \\ \hline
MISC \cite{BoXiao2012AnalyzingTL}                                 & MI on substance use by college students & MISC 2.1                                                                                                                 & text                         & 7293        & no                         \\ \hline
MITI  \cite{BoXiao2012AnalyzingTL}                                & MI on substance use by college students & MITI 2.0                                                                                                                 & text                         & 88          & no                         \\ \hline
\citeauthor{JamesGibson2016ADL} \cite{JamesGibson2016ADL}    & Counseling on substance abuse           & MITI 2.0, MISC 1.1                                                                                                       & text                         & 337         & no                         \\ \hline
CTT \cite{xiao2015rate}                                           & Counseling on substance abuse           & MITI 3.0                                                                                                                 & text, audio                  & 200         & no                         \\ \hline
\citeauthor{VernicaPrezRosas2017UnderstandingAP} \cite{VernicaPrezRosas2017UnderstandingAP} & Counseling therapy                      & MITI 4.1                                                                                                                 & text, audio                  & 276         & no                         \\ \hline
\citeauthor{AshishSharma2020ACA} \cite{AshishSharma2020ACA} & Online peer mental health support  & \begin{tabular}[c]{@{}l@{}}Emotional Reactions \\ Interpretations \\ Explorations\end{tabular}                         & text                         & 10,143      & yes                        \\ \hline
\textbf{MEDIC}                                                     & \textbf{Psychological counseling}       & \textbf{\begin{tabular}[c]{@{}l@{}}Expression of experience \\ Emotional reactions \\ Cognitive reaction\end{tabular}} & \textbf{text, audio, visual} & 771         & \textbf{yes}               \\ \hline
\end{tabular}}
\caption{Empathy datasets in the field of psychotherapy. }
\label{table:datasets}
\end{table*}

In this paper, we build a multimodal empathy dataset (MEDIC) for face-to-face counseling scenarios. To describe empathic communication in multimodal scenarios, we also propose three labels: expression of experience (EE), emotional reaction (ER) and, cognitive reaction (CR). EE describes whether the client has expressed experiences triggering empathy. ER and CR indicate the counselor’s empathic reactions
on emotion and cognitive dimensions, respectively. The dataset is constructed from counseling case videos. 
It contains textual modality about the content of the conversations, visual modality  about the body and face, and audio modality  about the voices. 

We evaluated the reliability of annotators' subjective evaluations by calculating the intraclass correlation coefficients and Fleiss' Kappa. Statistical analysis of these annotations showed that professional counselors tend to demonstrate cognitive empathy rather than emotional empathy when establishing empathic connections with their clients, particularly in psychotherapy scenarios. Furthermore, we constructed three baseline multimodal analysis models using the MEDIC dataset and found that each modality plays a significant role.



\section{Related Work}
Empathy has been researched due to its potential in psychology. There are two data-driven tasks: empathy prediction and empathy conversation generation. For empathy prediction, Alam et al. \cite{alam2018annotating} use speech data from call centers to predict whether agents would show empathy when facing anger or frustration from clients. Buechel et al. \cite{buechel2018modeling} predicte empathy when people read news reports. Barros et al. \cite{barros2019omg} asked people to tell stories and then captured the empathy of the listeners for empathy prediction. For empathy conversation generation, the  EMPATHETICDIALOGUES dataset contains 25,000 individual conversations based on specific contexts \cite{HannahRashkin2019TowardsEO}. A dialogue generation model is constructed to
generate empathic responses. 
Although empathy has been applied in the above domains, its computational approach has not been sufficiently studied in the field of psychotherapy due to a lack of datasets. Table \ref{table:datasets} shows existing data sets relevant to empathy in a psychotherapeutic context. They are described in detail in the following paragraphs.

 In the field of psychotherapy, empathy is generally used in the prediction task. Several empathy datasets are built by collecting texts. For example, a study from a clinical trial used motivational interactions (MI) to construct a dataset from therapy related to substance use in college students \cite{BoXiao2012AnalyzingTL}. Researchers manually transcribed the text from the therapist portion of the session. Then, they analyzed 28 MI sessions using the Motivational Interviewing Skill Code  (MISC) version 2.1 \cite{miller2003manual}, yielding 854 empathic and 6439 non-empathic utterances. The MISC manual describes the behavior of the counselor and client at the utterance-level and assesses the overall competence of the counselor. This dataset is named MISC. Then they evaluated the empathy level of 88 additional sessions using the Motivational Interviewing Treatment Integrity (MITI) version 2.0 \cite{moyers2003motivational}. The MITI is a session-level counselor coding scheme used to give a global score of counselor empathy on the Likert scale. This dataset is named MITI. The data used in the annotation of these two datasets includes audio and original transcripts. James Gibson et al. \cite{JamesGibson2016ADL} collected 337 texts from motivational interviews from six separate clinical studies. These studies all focused on addiction counseling related to substance abuse. The researchers manually transcribed and segmented all data at the turn level, and coded them for session-level behavior according to the MITI 2.0 manual. Then talk turns were segmented into  utterances, which were assigned utterance-level behavioral codes according to the MISC 1.1 manual \cite{PaulCAmrhein2008ManualFT}. Ashish Sharma et al. \cite{AshishSharma2020ACA} used two online mental health support platforms, TalkLife and  Mental Health Subreddits, as data sources to obtain 10,143 pairs of communicative texts. They devised a new method to describe empathy that incorporates emotional reaction, interpretation, and exploration. The degree of empathy depends on whether or not these responses are expressed. 

Texts are not the only vehicle for perceiving empathy. Pitch et al. \cite{reich2014vocal} carries information about the emotional state of the speaker and has been shown to be related to the perception of empathy in psychotherapy. Bao Xiao et al \cite{xiao2015rate}. further provided empathy prediction data, including text and audio, in Context Tailored Training (CTT). The dataset contains 200 motivational interviews, includes an observer rating of a counselor's empathy using MITI 3.0 \cite{moyers2007revised}. Sessions were transcribed using Automatic Speech Recognition (ASR), and the generated text was used in a text-based empathy prediction model. These speeches and language processing techniques could accurately predict therapists’ performance from the audio recordings. Another study used 276 MI audio sessions from clinical studies \cite{VernicaPrezRosas2017UnderstandingAP}. The full set consists of 97.8 hours of audio, with an average session length of 20.8 minutes. The researchers used MITI 4.1 \cite{TBMoyers2022MotivationalIT} to label the client's and counselor's empathy levels separately.

All of the above datasets contain only text and/or audio modalities. However, empathy can also be expressed visually. For example, facial expressions and gaze can be used to infer empathic behavior \cite{kumano2011analyzing}. Therefore, we introduce visual modality into empathy research in the field of counseling. In addition, most of the above datasets use different versions of the MISC and MITI scales to measure empathy. These are two classic empathy measurement scales, but neither of them takes into accounts for the complexity of empathy \cite{AshishSharma2020ACA}. They assess only cognitive empathy and ignore emotional empathy. More importantly, MISC describes client and counselor behavior at the utterance-level, while MITI describes counselor behavior at the session-level \cite{BoXiao2012AnalyzingTL}. This does not match the talk turn level sample on our dataset, so we do not use either of these scales. Sharma et al.’s \cite{AshishSharma2020ACA} dataset considers both cognitive and emotional empathy. Their description of cognitive empathy in terms of both interpretation and exploration is illuminating, but they only consider the strength of expressions and ignore correctness. They may misinterpret cognitive expressions as highly emotional ones. Therefore, we have combined the above empathy scales for our annotation. We consider both emotional and cognitive empathy. To fit the talk turn level sample, we also introduce an empathy cycle in the labels \cite{barrett1981empathy}. Finally, only the dataset from online psychological support platforms with non-professional counseling is publicly available. However, we will make our dataset available for future research.

In contrast to the existing datasets, our proposed dataset, MEDIC, introduces visual modality into empathy research in the field of counseling. It considers both emotional and cognitive empathy and is designed to fit the talk turn level sample. Furthermore, our dataset will be publicly available for easy access, making it a valuable resource for empathy research in the field of psychotherapy.

\begin{figure*}[t]
\centering
\includegraphics[width=0.97\textwidth]{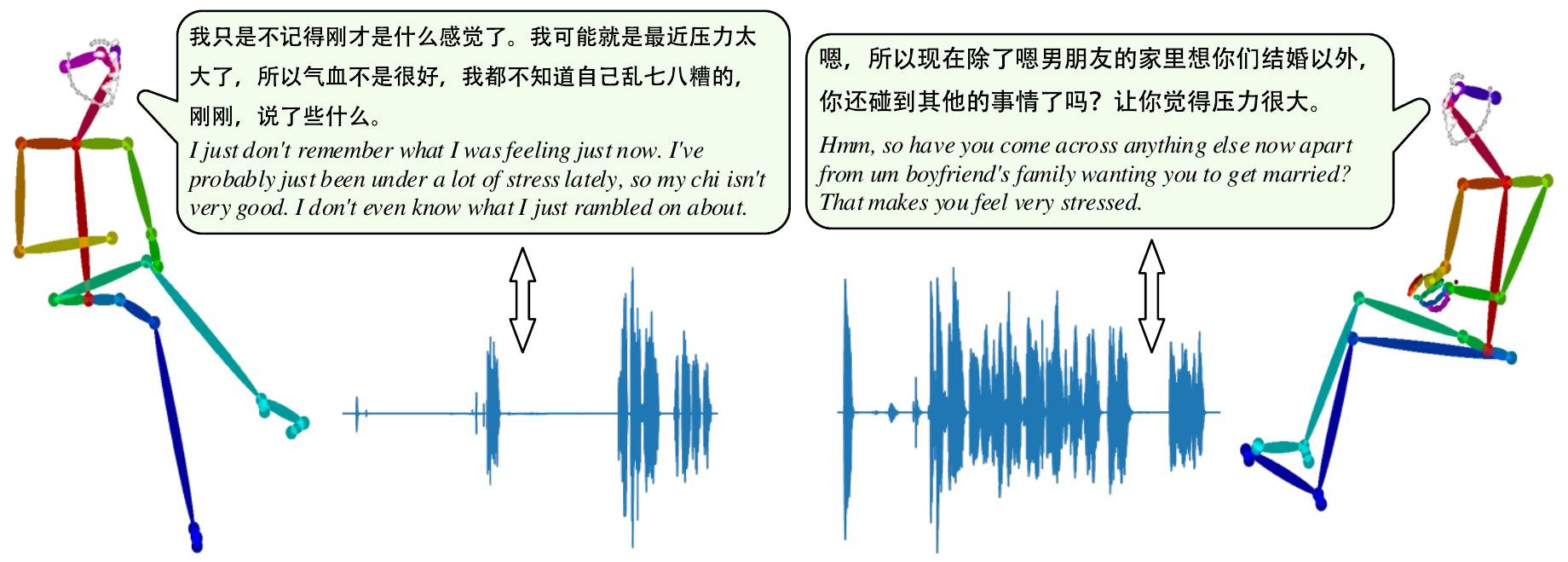} 
\caption{The sample of our dataset holds three modes of data: visual, textual and audio. The italicized text is the English translation of the original. On the left, the client is expressing his or her situation. On the right the counselor is asking about the reasons for the client's emotion. }
\label{fig:sample}
\end{figure*}

The major contributions of this paper are:
\begin{itemize}
\item To the best of our knowledge, MEDIC is the first empathy dataset for psychotherapy scenarios that  considers visual, audio, and textual modalities. 
\item We devise a new method for describing empathic communication in multimodal scenarios.
\item We construct three multimodal analysis baseline models on the MEDIC dataset to demonstrate the contribution of each modality in predicting psychological empathy. 
\end{itemize}

\section{How to Measure Empathy}
Client-centered therapy and psychoanalytics are two therapeutic approaches to empathy in psychotherapy corresponding to the cognitive and emotional aspects of empathy, respectively \cite{selman1980growth}.  Some traditional counselors emphasize a conscious perspective selection process rather than a more automatic body-based emotion simulation process \cite{bohart1997empathy}. However, empathy is a complex multidimensional concept, encompassing the individual's perceptions, transpersonal abilities or dispositions, and emotional reactions \cite{MatthewDavis1980AMA}. Empathy research has also been carried out in emotional \cite{HannahRashkin2019TowardsEO} and cognitive \cite{xiao2015rate} directions.

One empathic process model, the Empathy Cycle \cite{barrett1981empathy}, is particularly relevant. In the Empathy Cycle, the client and therapist work together to find an accurate expression of the client's experience. The cycle includes four steps. First, the client expresses the experience. The counselor generates empathy, and then expresses it. Lastly, the client accepts empathy. 
Counselor-generated empathy and client-accepted empathy are inherently unobservable. 


Inspired by these studies, we propose a new conceptual method for describing empathy. This method consists of three mechanisms of empathy communication. We find the Empathy Cycle fits perfectly with our talk turn level sample. Therefore, we choose client expression of experience and counselor expression of empathy as important components of our scale. In conjunction with the multidimensional meaning of empathy mentioned above, we subdivide counselor-expressed empathy into cognitive and emotional reactions. In total, this method consists of three mechanisms of empathy communication: expression of experience, emotional reaction, and cognitive reaction. For each mechanism, we adopt a three-category scale: no expression (0), weak expression (1), and strong expression (2). 
The method is described in detail below. 

\textbf{Expression of Experience (EE). } The expression of the client's experience is the first step in the empathy cycle. Empathy is an internal experience; responsive empathy corresponds to the responder \cite{barrett1981empathy}. The responder cannot express empathy to an experience that has not been expressed in some way. The expressing effect depends on the quality of the receiver, the signal, and the sender, so the signal must be measured. The experience here includes not only what clients have done, but also how they feel. Therefore, they should ideally express their feeling or describe an experience. When EE is no expression (0), there are no expressed emotions or described experiences. EE with weak expression expresses a weak emotion or mentions an experience. EE with strong expression corresponds to a strong emotion or a full description of an experience. 

\textbf{Emotional Reactions (ER). } The emotional reaction expressed by the counselor is part of the empathic reaction. The counselor first observes the expression of the client's experience, then develops and expresses empathy.  The counselor usually expresses emotions such as warmth and compassion. Different modalities have different forms of presentation for emotional reactions. The text contains explicit and/or implicit emotional words \cite{buechel2018modeling}. 
In addition to the verbal information, various acoustic features in audio, such as pitch and loudness, also contain rich emotional reactions \cite{alam2018annotating}. The emotional reactions contained in the videos are primarily reflected in facial expressions and body movements \cite{barros2019omg}. For generalisability reasons, the emotional reaction labels should correspond to no emotional reaction, weak emotional reaction, and strong emotional reaction. For example, in everyday conversation, when a person shares a very funny experience, the listener may laugh out loud. However,  we found that in our dataset, the counselor did not express strong emotional reactions. The final labels in the dataset for emotional reactions include only no emotional reactions (0) and weak emotional reactions (1).

\begin{table*}[t]
\centering
\begin{tabular}{c|ccc|l|l}
\hline
\multicolumn{1}{l|}{} & \multicolumn{1}{l}{EE} & \multicolumn{1}{l}{ER} & \multicolumn{1}{l|}{CR} & Talker    & Message                                                                                                                                                                                               \\ \hline
\multirow{2}{*}{1)}   & \multirow{2}{*}{2}     & \multirow{2}{*}{1}     & \multirow{2}{*}{2}      & Client    & Did I do a bad job? Did I do something wrong?                                                                                                                                                         \\ \cline{5-6} 
                      &                        &                        &                         & Counselor & \begin{tabular}[c]{@{}l@{}}It doesn't seem like your daughter has been affected by you telling her about \\ this at all yet. On the contrary, she feels that she can comfort her mother. \end{tabular} \\ \hline
\multirow{5}{*}{2)}   & \multirow{5}{*}{1}     & \multirow{5}{*}{0}     & \multirow{5}{*}{0}      & Client    & Huh? No more? So soon today?                                                                                                                                                                          \\ \cline{5-6} 
                      &                        &                        &                         & Counselor & Hmm.                                                                                                                                                                                                  \\ \cline{5-6} 
                      &                        &                        &                         & Client    & Well, good.                                                                                                                                                                                           \\ \cline{5-6} 
                      &                        &                        &                         & Counselor & We can continue next week.                                                                                                                                                                            \\ \cline{5-6} 
                      &                        &                        &                         & Client    & Good.                                                                                                                                                                                                 \\ \hline
\multirow{2}{*}{3)}   & \multirow{2}{*}{0}     & \multirow{2}{*}{0}     & \multirow{2}{*}{2}      & Client    & What should I say?                                                                                                                                           \\ \cline{5-6} 
                      &                        &                        &                         & Counselor & You can start with when you and your boyfriend met.                                                                                                                                               \\ \hline
\end{tabular}
\caption{Typical examples of three extreme cases. These samples contain labels with very high and very low values. Due to space limitations, only the translations of the text content and the corresponding labels are provided here. }
\label{table:samples}
\end{table*}

\textbf{Cognitive Reactions (CR). } The cognitive reaction expressed by the counselor is another part of the empathic reaction. 
Sharma et al.  suggest that cognitive reactions can be divided into interpretation and exploration \cite{AshishSharma2020ACA}. No cognitive reaction means that the counselor does not refer to or further explore the client's feelings and experiences. A weak cognitive reaction means that the counselor mentions or expresses interest in the client's feelings and experiences. The feelings and experiences may be not accurate. A strong cognitive reaction means that the counselor explicitly explains or explores the client's feelings and experiences. This explanation must accurately correspond to the client's feelings or experiences.

\section{Data Collection}
\subsection{Data Source}
Our data is based on UM Psychology's counseling case videos\footnote{https://appdu96wh3o1781.h5.xiaoeknow.com/v1/goods/good
s\_detail/p\_5fd6221ce4b04db7c094a079?type=3\&type=3\&jump\_f
rom=1\_05\_37\_01}. The case videos contain 11 hours of recorded interactions between clients and counselors. The cases cover counseling issues including marital relationships, professional dilemmas, family education, the meaning of life, and other topics. The cases also contain the reactions of counselors while dealing with clients with a variety of attitudes. Both the client and the counselor in the video are UM Psychology counselors, and they acted out some cases that used to be real. Each case is played out by two different counselors. The client and counselor sit face to face and a camera records the movements and expressions of both. The 10 cases are divided into 38 videos, each averaging 17 minutes and 50 seconds. In total, we record 678 minutes (approximately 11 hours) of video. 

\subsection{Data Pre-Processing}
For privacy reasons, we manually removed personal information and cropped out any mentions of specific people or places. For the image data, we use OpenPose\cite{cao2017realtime} to extract feature points from the face, torso, and arms. The final data set will use feature points to protect user privacy. As our data source consists of a re-enactment of the teaching video, the original patient is absent from the visual content. This approach effectively mitigates potential ethical issues that may arise from the use of identifiable patient data. These ensure that our data set can be public.

Counseling videos are cropped according to talk turns. According to the Empathy Cycle, the dialogue rounds start with the client expressing an experience and end with the counselor expressing an empathic reaction (presumably accepted by the client). 
However, dialogue is frequently interrupted by vocalized pauses (e.g., uh, um). One of the more common scenarios for this is when the counselor uses “um” during the conversation to express that he or she is listening intently to what the other person is saying. This is a trick or habit in conversation. However, a problem arises when these pauses are used as segments. In such cases, the counselor's words do not convey any meaningful information, making it difficult for the ER and CR to make accurate judgments due to the lack of information. These vocalized pauses are not used to split the talk turn, so an overall talk turn starts with content expressed by the client and ends with the counselor. Each talk turn corresponds to a sample and has a set of labels. As the talk turn contains both client and counselor sections, it can be further broken down by person. Finally, we separated the audio and image data and used an audio text recognition tool to extract the approximate text modality. The audio-to-text tool we use is a video and audio editing tool from Wondershare\footnote{https://uniconverter.wondershare.cn/}. To ensure the accuracy of the text, the video and text are put back together for manual proofreading. Figure \ref{fig:sample} shows the form of our sample and contains visual modality about the client and the consultant, the respective audio, and the text of the conversation.

\subsection{Data Annotation}
Empathy is not a universal concept, and the meaning of empathy varies by culture and language. To ensure that linguistic meanings were understood accurately, data sources were annotated by five students who spoke the same language (Chinese). Annotators were thoroughly trained in our labeling methods and had to pass a live data annotation trial. Our annotators received full annotation instructions and some classic examples. We then conducted three one-hour offline sessions to explain details about empathy and each label’s meaning. During this process, annotators were asked questions about annotations, which largely resolved potential ambiguity issues. Before formal annotation began, we conducted an on-site test to ensure that each annotator was qualified for annotation. 

The annotator pairs watched video clips of each sample and identified three mechanisms (the client's expression of experience and the counselor's emotional and cognitive reactions). 
During the annotation process, the annotator is asked to pay attention not only to the information contained in the spoken words but also to the voice and the intonation. At the same time, they must consider the expressions and actions of clients and counselors. Each sample was annotated by two annotators. For each label, we verify if the two annotators’ results are identical. If there is a discrepancy in the results, a third annotator will review the video clip again to determine which result is superior.

\begin{table}[t]
\centering
\begin{tabular}{c|c|c}
\hline
                    & \multicolumn{1}{c|}{Average Sample}   & \multicolumn{1}{c}{Total} \\ \hline
Talk Turns          & 1                                     & 771                       \\ \hline
Speaking Times      & 4.29                                  & 3306                      \\ \hline
Number of Words     & 129.45                                & 99808                     \\ \hline
Duration            & 52.76s                                & 678m                      \\ \hline
Number of Frames    & 1137                                  & 876692                    \\ \hline
\end{tabular}
\caption{ Dataset statistics. }
\label{table:Statistics}
\end{table}

\begin{figure}[t]
\centering
\includegraphics[width=.9\columnwidth]{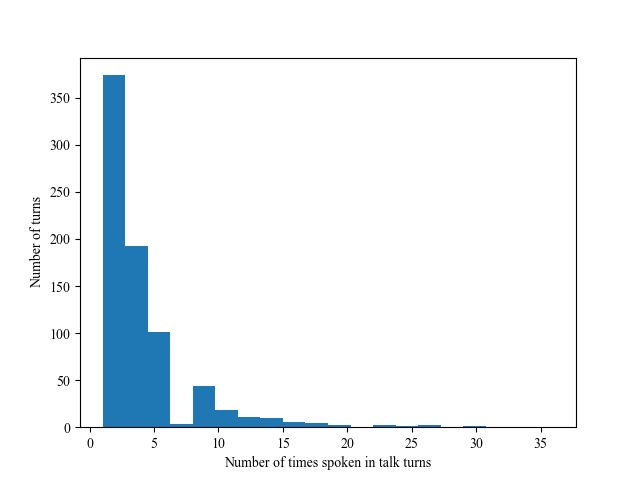} 
\caption{Distribution of times spoken in talk turns. }
\label{fig:spoken}
\end{figure}

\subsection{Sample Analysis}
Table \ref{table:samples} shows a few typical samples 
on all three labels. This means that the client engages in strong expression, and the counselor gives a relatively strong empathic reaction. The client expresses doubt about her's own behavior and shows a strong desire for approval. In addition, although no evidence is given in the forms, the client shows an expression of anxiety in the images. The client's voice sounds anxious. Therefore, in this sample, the EE was marked as expressing strong emotions. Based on the context, the client is concerned that her behavior has a negative effect on her daughter. The counselor uses a soft voice and gentle expression to reassure the client that her daughter is not affected. The ER label corresponds to a weak emotional reaction. Finally, from the cognitive aspect, the counselor clearly explains the client's previous behavior and further reassures the client. Therefore, the CR label corresponds to a strong cognitive reaction. 

The second and third samples received lower scores on the EE, ER, and CR labels. The lowest scores were obtained on the second ER and CR labels. The client was a little surprised at the end of the consultation and expressed weak emotions, so EE was annotated as a weak expression. The counselor had only a simple reaction to the client’s words and did not react emotionally or cognitively. Therefore, ER and CR are marked as having no empathic reaction. The third sample had the lowest EE and ER scores. The client initially asks what she should say, and no emotion or experience is expressed. Therefore, EE corresponds to no expression. The counselor leads the client to discuss past experiences. Therefore, CR corresponds to strong cognitive reactions. ER is marked as no expression because the counselor does not express emotional reaction.

\begin{figure}[t]
\centering
\includegraphics[width=.9\columnwidth]{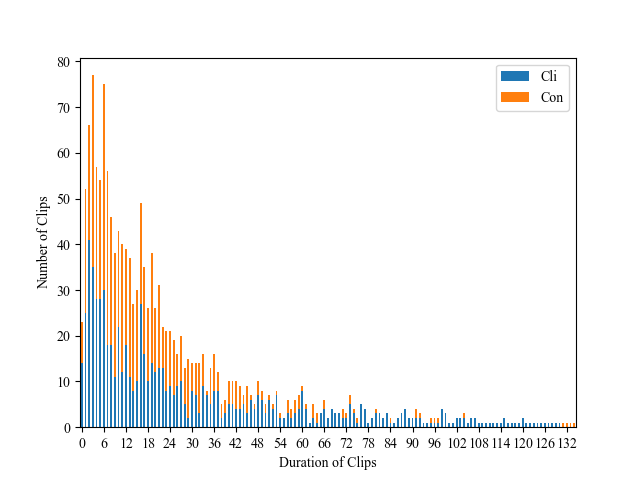} 
\caption{Distribution of duration of single speaking.}
\label{fig:duration}
\end{figure}

\begin{figure}[t]
\centering
\includegraphics[width=.9\columnwidth]{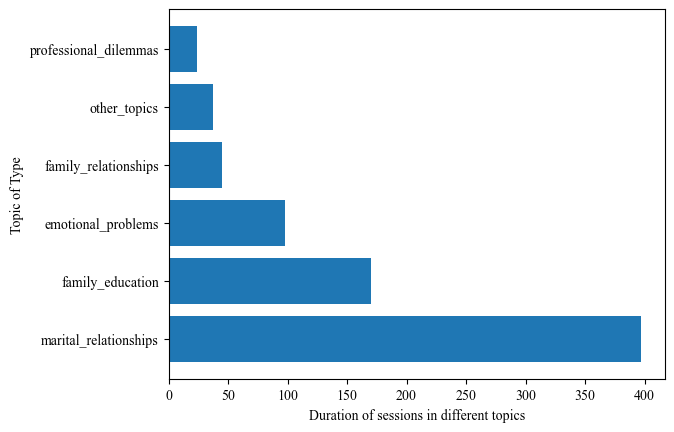} 
\caption{Duration of sessions in different topics }
\label{fig:topic_duration}
\end{figure}

\subsection{Statistical Results}

Individual statistical results are given in Table \ref{table:Statistics}. The final dataset contains 771 talk turns, each corresponding to one sample. The average length of each talk turn is 53 seconds. Each talk turn contains approximately 4 sentences, 53 seconds of audio, and 1137 frames of image features. 

\begin{figure*}[!htbp]
	\centering
	\subfigure{
		\begin{minipage}[t]{0.32\linewidth}
			\centering
			\includegraphics[width=1\textwidth]{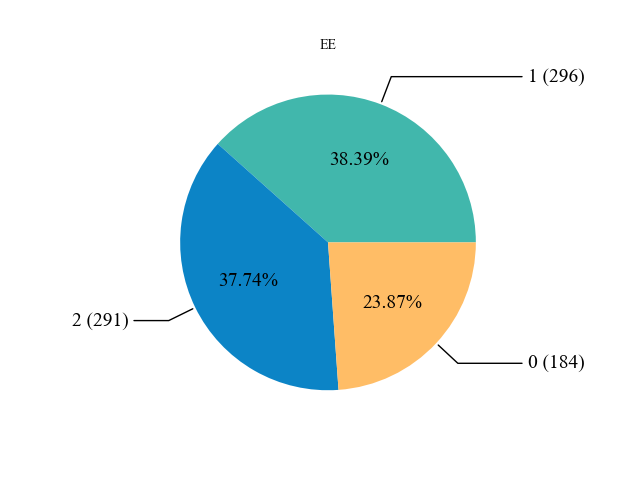}
		\end{minipage}%
	}%
	\subfigure{
		\begin{minipage}[t]{0.32\linewidth}
			\centering
			\includegraphics[width=1\textwidth]{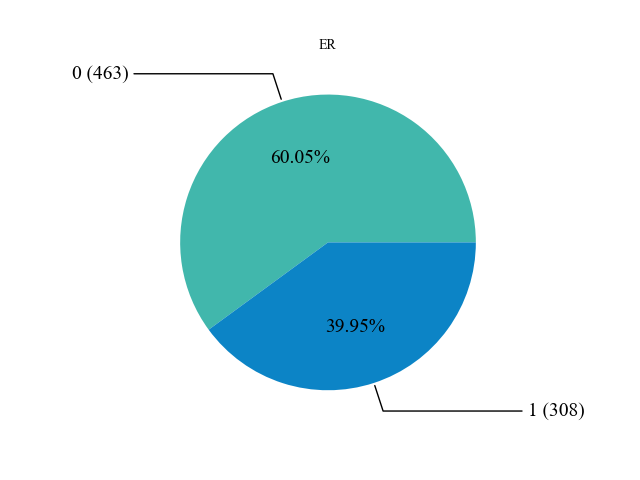}
		\end{minipage}
	}%
	\subfigure{
		\begin{minipage}[t]{0.32\linewidth}
			\centering
			\includegraphics[width=1\textwidth]{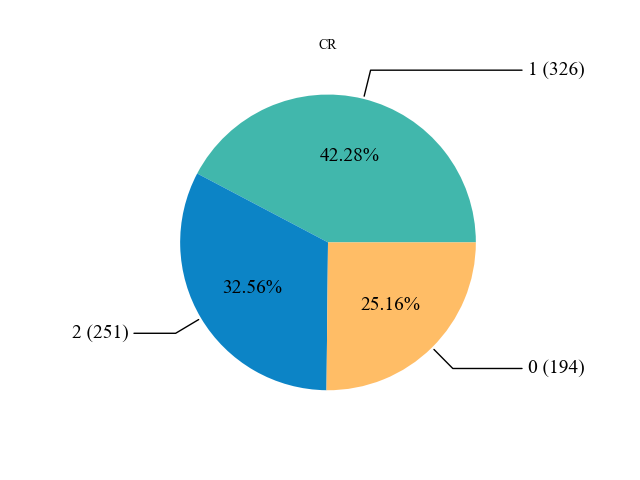}
		\end{minipage}
	}
	\centering
	\caption{The overall distribution of EE, ER, and CR scores.}\label{fig:label}
\end{figure*}

Figure \ref{fig:spoken} shows the distribution of the number of times spoken. The majority of talk turns contain only a few exchanges, with client and counselor typically only speaking once each. In some talk turns, the client or consultant uses a vocalized pause as a reaction. We do not split the talk round at the pause. On average, each client or counselor speaks 4.29 times in a talk turn, for a total of 3306 times. The distribution of the length of each person's speech is shown in Figure \ref{fig:duration}. In most cases it is very short. However, it takes longer for the client to describe personal experiences. In addition, the average length of a single client comment is 11.48s, much more than the length of a single counselor comment of 6.59s. The graph also shows that counselors speak longer when the conversation is very short, and the clients speak more when the conversation is very long. In counseling, clients spend most of the time describing experiences while the counselor listens. Figure \ref{fig:topic_duration} presents the total duration of consultations across different themes. The consultations on marital relationships lasted the longest and were significantly longer than the others. This suggests that marital relationships have a major influence on people’s mental health and well-being, and that many people face challenges and hardships in their marriages and require professional assistance and advice. Among all the themes, only professional dilemmas are solely related to their own issues, while most of the other themes stem from relationships with others. This also indicates that interpersonal relationships are one of the key factors affecting mental health.

Figure \ref{fig:label} shows the distribution of labels for the final samples. Overall the distribution of all three labels is relatively balanced. For clients, the percentage of EE labeled 0 is only 24\%, indicating that clients are comfortable confiding about their experiences to the counselor. 
For the counselor, the unrecognized expression of CR is only 25\%. This indicates that counselors tend to use cognitive empathy to interact with the client. In contrast, no emotional expressions account for 60\% of ER, even when strong emotional expressions are removed beforehand. This indicates that counselors tend not to express empathy through emotions. This situation proves that modern counselors focus on cognitive expression rather than emotional expression, which is consistent with previous research \cite{bohart1997empathy}.

\begin{figure}[t]
\centering
\includegraphics[width=.9\columnwidth]{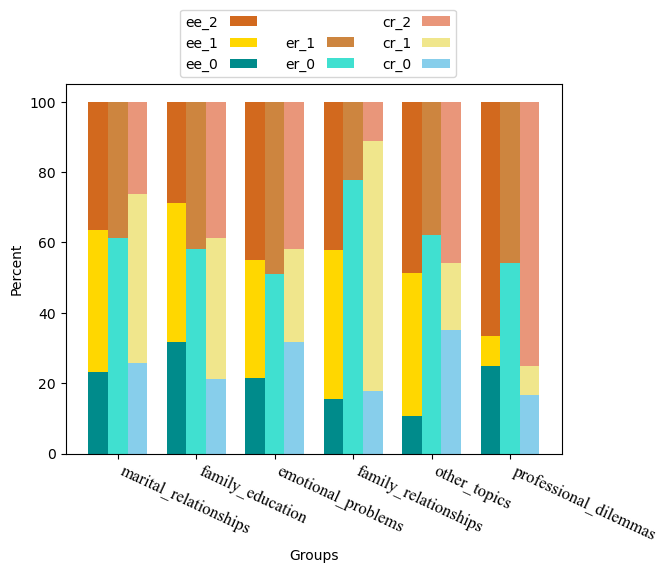} 
\caption{Distribution of labels under different topics }
\label{fig:topic_label}
\end{figure}

Figure \ref{fig:topic_label} shows the distribution of labels under different topics. The distribution of Expression of Experience and counselor responses is relatively similar across the different topics, except for family relations and career dilemmas. The topics related to marital relationships, family education, and emotional problems have a more balanced distribution of labels. For family relations, the counselor mostly gave weak responses. For career dilemmas, both the client’s Expression of Experience and the counselor’s responses were strong. This may suggest that the counselor views career difficulties as more personal than situational or interpersonal, and thus tries to help the clients recognize their own problems and resources and improve their problem-solving skills.

Pearson correlation coefficients of r = 0.45 are calculated for the two labels ER and CR. It indicates that the two scales are somewhat correlated but capture different empathy phenomena, emphasizing the importance of using multiple variables to describe the empathic reaction. In terms of label reliability, we analyzed consistency between annotators for each label. Each sample was scored by two labelers and only given to a third for decision-making if the first two did not agree. We measured annotation consistency between the two individuals using the intraclass correlation coefficient (ICC) \cite{koo2016guideline}, Fleiss’ Kappa \cite{fleiss1971measuring}, and percentage agreement. The results are shown in Figure \ref{table: Consistency}.The values of Fleiss’ Kappa range from -1 to 1, where 0.61-0.80 signifies significant agreement. For ICC, values range from 0 to 1, with values closer to 1 indicating greater reliability. Values less than 0.5 indicate poor reliability, between 0.5 and 0.75 indicate moderate reliability, and between 0.75 and 0.9 indicate good reliability. The values of Percent Agreement range from 0 to 1, where 0 means complete disagreement and 1 means complete agreement among evaluators. This means all three labels in the dataset have a good confidence level. For consistency percentages, more than half of the annotated results for each label were free from ambiguity. Our dataset shows high reliability by integrating all results from the three metrics mentioned above.


\section{Evaluation Experiments}
To evaluate our dataset, features are extracted from three modalities and empathy classification is performed on three different models. TFN \cite{AmirZadeh2017TensorFN} is a highly cited classical model, and the SWAFN \cite{MinpingChen2020SWAFNSW} model focuses on textual modality, which is of paramount importance for consulting. A simple concatenation model is used as a comparison.

\begin{table}[t]
\centering
\begin{tabular}{l|l|l|l}
\hline
Measure           & EE     & ER     & CR     \\ \hline
Fleiss' Kappa     & 0.7557 & 0.6991 & 0.7102 \\ \hline
ICC               & 0.9021 & 0.8126 & 0.8732 \\ \hline
Percent Agreement & 0.8067 & 0.8586 & 0.7846 \\ \hline
\end{tabular}
\caption{ Consistency of annotation across the three labels. }
\label{table: Consistency}
\end{table}

\begin{figure}[ht]
\centering
\includegraphics[width=1\columnwidth]{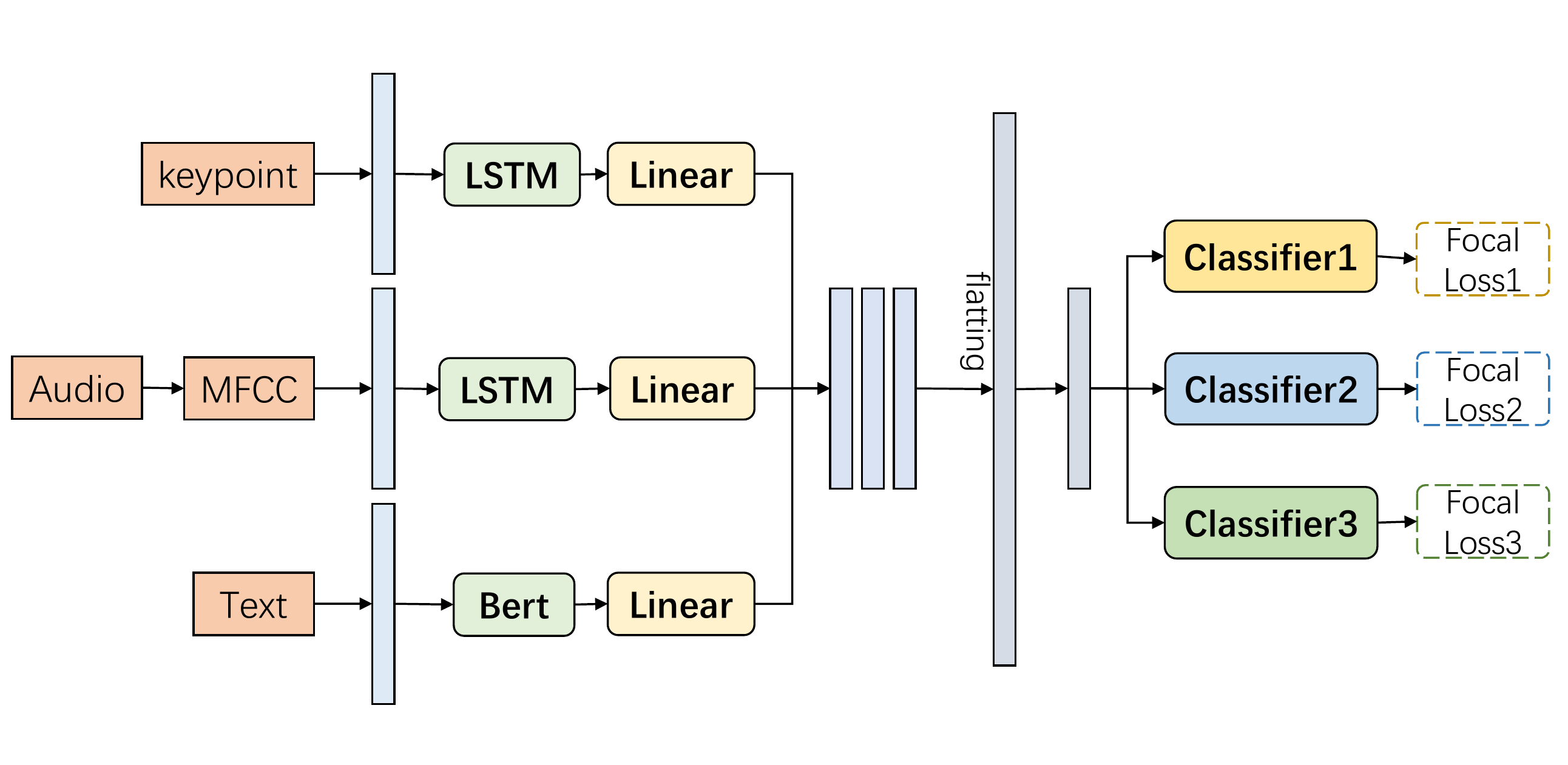} 
\caption{A simple framework for empathy prediction.}
\label{fig:model}
\end{figure}

\subsection{Feature Extraction}
\textbf{Visual features ($v$). } As mentioned previously, OpenPose \cite{cao2017realtime} was used to extract keypoints of the face, hands, and body for each frame of the video for the client and consultant. The keypoint coordinates were then normalized. A distinction was also made between the keypoints of the client and the consultant. The final input image feature dimension was $d_v \in R^{T_v \times 411}$, where $T_v$ is the number of frames in the clip. 

\textbf{Audio features ($a$). } Audio features corresponding to the segments were extracted using librosa \cite{mcfee2015librosa}. The audio is separate for the client and the consultant. In this experiment, MFCC features are used for training. The final input audio feature dimension is $d_a \in R^{T_a \times 20}$, where $T_a$ is determined by the duration of the audio. 

\textbf{Text features ($t$). } Our text is derived from a transcription. We use the Bert pre-training model to extract text features. The dimension of the final input text feature is $d_t \in R^{T_t \times 768}$, where $T_t$ is determined by the length of the sentence. 

\subsection{Baseline Model}

\textbf{The TFN model. } TFN \cite{AmirZadeh2017TensorFN} consists of three components. The sub-network takes single-modal features as input and a multimodal embedding as output. A tensor fusion layer uses a 3-fold Cartesian product to explicitly model unimodal, bimodal, and trimodal interactions. A neural network uses the output of the tensor fusion layer as input to predict empathy. 

\textbf{The SWAFN model. } SWAFN \cite{MinpingChen2020SWAFNSW} uses a co-attentive mechanism to learn bidirectional large-scale contextual information between  language and other modalities. The sentiment word classification task is then integrated into the model via a multi-task learning mechanism that guides the learning and aggregation of multimodal fusions.

\textbf{The concatenation model. } We also use a very simple multimodal concatenation model as shown in Figure \ref{fig:model}. The temporal information of the three modalities are extracted using LSTM. Then they are spliced together and connected to the classifier to predict empathy.

Since the three labels are derived from two different people, in this paper we separately predict corresponding labels using data from the client and the counselor. The client's features are used to predict EE, and the counselor's features are used to predict ER and CR.

\subsection{Experiment setup}

\textbf{Definition.} Let $X_x \in \mathbb{R}^{\tau_x \times d_x}$ denote the features of a modality. In our dataset, $x$ can represent visual ($v$), audio ($a$), and text ($t$) modalities. $\tau_x$ represents the temporal length of the modality, and $d_x$ represents the feature dimension of the modality. Let $Y = \{y_1, y_2,…, y_l\}$ denote the label space with $l$ labels. In our experiments for the three labels EE, ER, and CR, $l=3$. For the classification task of predicting EE, ER, and CR, the goal is to predict the corresponding label $y_i$ given the input feature of clip $X_{x_i}$, where $x \in \{a, v, t\}$.

\begin{table}[t]
\centering
\begin{tabular}{c|l|l|l|l}
\hline
\multicolumn{1}{l|}{}      & label & EE  & ER  & CR  \\ \hline
\multirow{3}{*}{train set} & 0     & 127 & 339 & 145 \\ \cline{2-5} 
                           & 1     & 204 & 200 & 234 \\ \cline{2-5} 
                           & 2     & 208 & *   & 160 \\ \hline
\multirow{3}{*}{val set}   & 0     & 26  & 39  & 20  \\ \cline{2-5} 
                           & 1     & 28  & 38  & 24  \\ \cline{2-5} 
                           & 2     & 23  & *   & 33  \\ \hline
\multirow{3}{*}{test set}  & 0     & 31  & 85  & 29  \\ \cline{2-5} 
                           & 1     & 64  & 70  & 68  \\ \cline{2-5} 
                           & 2     & 60  & *   & 58  \\ \hline
\end{tabular}
\caption{ The number of labels in each category in the divided dataset. }
\label{table:divided_label}
\end{table}

\textbf{Sample details.} We conduct experiments on our dataset, which contains 771 samples. Each sample contains key points of images, audio, and dialogue text from both the client and the consultant. Each sample is annotated with the corresponding EE, ER, and CR. The training, test, and validation sets are split in a ratio of 7:1:2. Details of each category within each label are shown in Table \ref{table:divided_label}. The evaluation metrics are average accuracy(Acc) and macro F1-score.

\subsection{Implementation Details}

The TFN and SWAFN were trained on the V100 GPU and the concatenation model was trained on the RTX 3090 GPU. The pytorch and Adam optimizer is used on all models. The labels are set with the weights of the classes, which are inversely proportional to the number of classes due to their unbalance. The learning rate in TFN is set to 1e-4, the batch size is set to 32, and the dropout is set to 0.3. 
The learning rate in SWAFN is set to 1e-3, the batch size is set to 32, and the dropout is set to 0.3. 
The learning rate in the concatenation model is set to 1e-4, the batch size is set to 16, and the dropout is set to 0.4. 

\begin{table}[t]
\centering
\setlength{\tabcolsep}{1.5mm}
\begin{tabular}{l|ll|ll|ll}
\hline
\multicolumn{1}{c|}{\multirow{2}{*}{Modal}} & \multicolumn{2}{c|}{TFN}                                      & \multicolumn{2}{c|}{SWAFN}                                    & \multicolumn{2}{c}{Concatenation}                            \\ \cline{2-7} 
\multicolumn{1}{c|}{}                       & \multicolumn{1}{c|}{Acc}            & \multicolumn{1}{c|}{F1} & \multicolumn{1}{c|}{Acc}            & \multicolumn{1}{c|}{F1} & \multicolumn{1}{c|}{Acc}            & \multicolumn{1}{c}{F1} \\ \hline
v+a+t                                       & \multicolumn{1}{l|}{\textbf{0.758}} & \textbf{0.758}          & \multicolumn{1}{l|}{\textbf{0.864}} & \textbf{0.863}          & \multicolumn{1}{l|}{\textbf{0.819}} & \textbf{0.810}         \\ \hline
v+a                                         & \multicolumn{1}{l|}{0.555}          & 0.526                   & \multicolumn{1}{l|}{0.721}          & 0.718                   & \multicolumn{1}{l|}{0.716}          & 0.708                  \\ \hline
v+t                                         & \multicolumn{1}{l|}{0.746}          & 0.744                   & \multicolumn{1}{l|}{0.805}          & 0.808                   & \multicolumn{1}{l|}{0.813}          & 0.804                  \\ \hline
a+t                                         & \multicolumn{1}{l|}{0.738}          & 0.736                   & \multicolumn{1}{l|}{0.851}          & 0.852                   & \multicolumn{1}{l|}{0.800}          & 0.774                  \\ \hline
v                                           & \multicolumn{1}{l|}{0.416}          & 0.307                   & \multicolumn{1}{l|}{0.636}          & 0.595                   & \multicolumn{1}{l|}{0.325}          & 0.301                  \\ \hline
a                                           & \multicolumn{1}{l|}{0.527}          & 0.465                   & \multicolumn{1}{l|}{0.695}          & 0.686                   & \multicolumn{1}{l|}{0.697}          & 0.699                  \\ \hline
t                                           & \multicolumn{1}{l|}{0.729}          & 0.726                   & \multicolumn{1}{l|}{0.857}          & 0.857                   & \multicolumn{1}{l|}{0.768}          & 0.763                  \\ \hline
\end{tabular}
\caption{Baseline and ablation experiments for EE. }
\label{table:EE}
\end{table}

\begin{table}[t]
\centering
\setlength{\tabcolsep}{1.5mm}
\begin{tabular}{l|ll|ll|ll}
\hline
\multicolumn{1}{c|}{\multirow{2}{*}{Modal}} & \multicolumn{2}{c|}{TFN}                                      & \multicolumn{2}{c|}{SWAFN}                                    & \multicolumn{2}{c}{Concatenation}                            \\ \cline{2-7} 
\multicolumn{1}{c|}{}                       & \multicolumn{1}{c|}{Acc}            & \multicolumn{1}{c|}{F1} & \multicolumn{1}{c|}{Acc}            & \multicolumn{1}{c|}{F1} & \multicolumn{1}{c|}{Acc}            & \multicolumn{1}{c}{F1} \\ \hline
v+a+t                                       & \multicolumn{1}{l|}{0.729}          & 0.719                   & \multicolumn{1}{l|}{0.743}          & 0.743                   & \multicolumn{1}{l|}{\textbf{0.703}} & 0.699                  \\ \hline
v+a                                         & \multicolumn{1}{l|}{0.646}          & 0.639                   & \multicolumn{1}{l|}{0.743}          & 0.744                   & \multicolumn{1}{l|}{0.639}          & 0.631                  \\ \hline
v+t                                         & \multicolumn{1}{l|}{0.725}          & 0.720                   & \multicolumn{1}{l|}{0.770}          & 0.761                   & \multicolumn{1}{l|}{0.697}          & \textbf{0.712}         \\ \hline
a+t                                         & \multicolumn{1}{l|}{\textbf{0.743}} & \textbf{0.734}          & \multicolumn{1}{l|}{\textbf{0.776}} & \textbf{0.777}          & \multicolumn{1}{l|}{0.690}          & 0.688                  \\ \hline
v                                           & \multicolumn{1}{l|}{0.600}          & 0.375                   & \multicolumn{1}{l|}{0.592}          & 0.598                   & \multicolumn{1}{l|}{0.568}          & 0.549                  \\ \hline
a                                           & \multicolumn{1}{l|}{0.642}          & 0.595                   & \multicolumn{1}{l|}{0.724}          & 0.727                   & \multicolumn{1}{l|}{0.652}          & 0.647                  \\ \hline
t                                           & \multicolumn{1}{l|}{0.734}          & 0.717                   & \multicolumn{1}{l|}{0.770}          & 0.764                   & \multicolumn{1}{l|}{0.658}          & 0.657                  \\ \hline
\end{tabular}
\caption{Baseline and ablation experiments for ER. }
\label{table:ER}
\end{table}

\subsection{Experimental Results and Analysis}
Tables \ref{table:EE}, \ref{table:ER}, and \ref{table:CR} show the experimental results of three models for the EE, ER, and CR of prediction tasks. The relevant ablation experiment about visual(v), audio (a), and text(t) modalities are also included. For example, $v+a+t$ denotes the fusion of the three modalities, while $v+a$ denotes the fusion of two modalities $v$ and $a$. From the tables, we can draw the following conclusions.

First, from the three tables, we can find that the F1 scores and accuracy for EE, ER, and CR are all above 69\%. This demonstrates that our dataset is highly effective in facilitating empathic prediction.

Secondly, the three modal fusions of $v+a+t$ have achieved the best results. The results of two modal fusions are slightly lower, and the results of one modality alone are the lowest. For example, in Table \ref{table:CR}, the best performing model is SWAFN. Its F1-score of $v+a+t$ is 7.6\% higher than the results of $a+t$ and 11.5\% higher than the results of the text modality alone. Its accuracy of $v+a+t$ is 7.2\% higher than $v+t$ and 9.2\% higher than the results of the text modality alone. This indicates that the provided multimodal information and fusion are effective.

Focusing on the $v$ modality, we find that the results of $v+a+t$ are higher than $a+t$. For example, the three modal F1-score and accuracy of the TFN model in Table \ref{table:EE} are both 2\% higher than $a+t$. The three modal F1-score and accuracy of the Concatenation model in Table \ref{table:CR} are both 4.4\% higher than $a+t$. This shows the importance of the video modality.

\begin{table}[t]
\centering
\setlength{\tabcolsep}{1.5mm}
\begin{tabular}{l|ll|ll|ll}
\hline
\multicolumn{1}{c|}{\multirow{2}{*}{Modal}} & \multicolumn{2}{c|}{TFN}                                      & \multicolumn{2}{c|}{SWAFN}                                    & \multicolumn{2}{c}{Concatenation}                            \\ \cline{2-7} 
\multicolumn{1}{c|}{}                       & \multicolumn{1}{c|}{Acc}            & \multicolumn{1}{c|}{F1} & \multicolumn{1}{c|}{Acc}            & \multicolumn{1}{c|}{F1} & \multicolumn{1}{c|}{Acc}            & \multicolumn{1}{c}{F1} \\ \hline
v+a+t                                       & \multicolumn{1}{l|}{0.722}          & 0.712                   & \multicolumn{1}{l|}{\textbf{0.783}} & \textbf{0.785}          & \multicolumn{1}{l|}{\textbf{0.735}} & \textbf{0.731}         \\ \hline
v+a                                         & \multicolumn{1}{l|}{0.501}          & 0.439                   & \multicolumn{1}{l|}{0.658}          & 0.657                   & \multicolumn{1}{l|}{0.613}          & 0.581                  \\ \hline
v+t                                         & \multicolumn{1}{l|}{\textbf{0.734}} & \textbf{0.725}          & \multicolumn{1}{l|}{0.678}          & 0.661                   & \multicolumn{1}{l|}{0.697}          & 0.701                  \\ \hline
a+t                                         & \multicolumn{1}{l|}{0.716}          & 0.711                   & \multicolumn{1}{l|}{0.711}          & 0.709                   & \multicolumn{1}{l|}{0.684}          & 0.687                  \\ \hline
v                                           & \multicolumn{1}{l|}{0.436}          & 0.259                   & \multicolumn{1}{l|}{0.467}          & 0.420                   & \multicolumn{1}{l|}{0.490}          & 0.450                  \\ \hline
a                                           & \multicolumn{1}{l|}{0.500}          & 0.395                   & \multicolumn{1}{l|}{0.605}          & 0.570                   & \multicolumn{1}{l|}{0.594}          & 0.556                  \\ \hline
t                                           & \multicolumn{1}{l|}{0.690}          & 0.685                   & \multicolumn{1}{l|}{0.691}          & 0.670                   & \multicolumn{1}{l|}{0.710}          & 0.722                  \\ \hline
\end{tabular}
\caption{Baseline and ablation experiments for CR. }
\label{table:CR}
\end{table}

Table \ref{table:ER} shows that $a+t$ achieves the best results on the TFN and SWAFN models, while $v+a+t$ and $v+t$ achieves the best results in concatenation model. This indicates that $v+a+t$ contributes to the training of the ER, but different models focus on different modal information when predicting the ER.

Finally, in the unimodal ablation experiments, $t$ performs the best. In Table \ref{table:EE}, for the SWAFN model, the F1-score and accuracy of $t$ are 17.1\% and 16.2\% higher than $a$. In Table \ref{table:ER}, for the SWAFN model, the F1-score and accuracy of $t$ are 3.7\% and 4.6\% higher than $a$. In Table \ref{table:CR}, for the concatenation model, the F1-score and accuracy of $t$ are 16.6\% and 11.6\% higher than $a$. This is because the text contains the richest verbal information, which is considered to be the most important element in counseling.

\subsection{Error Analysis}
Our study reveals that all three models incorrectly labeled 1 as 0 for the ER labels in some samples, likely due to imbalanced data distribution. This highlights the need to enhance the models' ability to handle unbalanced samples. Moreover, we discovered that the models tended to rely on sample length, leading to incorrect predictions for some shorter and longer samples across all three labels. This indicates a deficiency in the models' ability to extract semantic meaning.

\section{Conclusion}
Empathy is a critical component in human interaction and plays an important role in facilitating communication between clients and counselors. There is a paucity of multimodal empathy datasets in the field of psychology, and the corresponding empathy evaluation criteria are not well developed. In this paper, we construct a publicly available multimodal dataset for empathy. In order to understand empathy expressed in counseling exchanges, we propose three labels to describe empathic communication in multimodal scenarios. The dataset contains image modality, audio modality, and text modality from counseling scenarios. Collecting multimodal empathy datasets can be challenging due to the need to protect privacy while gathering data from multiple sources. Additionally, accurately measuring empathy, a complex and multidimensional concept, requires carefully designed assessment criteria. As a result, the amount of data we were able to collect was limited. Despite this, the high-quality data in our dataset holds significant value for empathy research. We hope that more empathy studies in the field of counseling will build on this foundation.


\bibliographystyle{ACM-Reference-Format}
\bibliography{sample-base}


\end{document}